\documentclass[conference]{IEEEtran}

\usepackage[cmex10]{amsmath}
\usepackage{amsthm}
\usepackage{amssymb}
\usepackage{mathrsfs}
\usepackage{graphicx}
\usepackage{float}
\usepackage{array}
\usepackage{epstopdf}
\usepackage{multirow}

\begin{document}

\title{\huge{A\MakeLowercase{d-}N\MakeLowercase{et}: A\MakeLowercase{udio}-V\MakeLowercase{isual} C\MakeLowercase{onvolutional} N\MakeLowercase{eural} N\MakeLowercase{etwork} f\MakeLowercase{or} A\MakeLowercase{dvertisement} D\MakeLowercase{etection}  I\MakeLowercase{n} V\MakeLowercase{ideos }}}

\author{Shervin Minaee$^*$, Imed Bouazizi$^{\dagger}$, Prakash Kolan$^{\dagger}$,  Hossein Najafzadeh$^{\dagger}$  \\
$^*$New York University
\\ $^{\dagger}$Samsung Research America\\ \\
}

\maketitle

\begin{abstract}
Personalized advertisement is a crucial task for many
of the online businesses and video broadcasters.
Many of today's broadcasters use the same commercial for all customers, but as one can imagine different viewers have different interests and it seems reasonable to have customized commercial for different group of people, chosen based on their demographic features, and history.
In this project, we propose a framework, which gets the broadcast videos, analyzes them, detects the commercial and replaces it with a more suitable commercial. 
We propose a two-stream audio-visual convolutional neural network, that one branch analyzes the visual information and the other one analyzes the audio information, and then the audio and visual embedding are fused together, and are used for commercial detection, and content categorization.
We show that using both the visual and audio content of the videos significantly improves the model performance for video analysis.
This network is trained on a dataset of more than 50k regular video and commercial shots, and achieved much better performance compared to the models based on hand-crafted features.
\end{abstract}

\IEEEpeerreviewmaketitle

\section{Introduction}
The amount of videos available on Internet is increasing dramatically, due to the availability of consumer electronics with high quality cameras.
Therefore it is crucial for video hosting websites (such as Youtube) to be able to better recognize, index, search, recommend and essentially make sense of the video. More importantly, all these tasks need to be done reasonably fast.
Traditionally, hand-crafted features were used followed by classifiers (or more generally predictors) to solve these problems.
Different video descriptors have been proposed over the years.
Some of them are based on the extension of the 2D image descriptors to 3D, such as SIFT-3D \cite{sift3d} and HOG3D \cite{hog3d}. 
And some specifically designed for video analysis, such as ActionBank for action recognition \cite{actionbank}, and spatio-temporal interest points (STIPs) (which is inspired by  Harris corner detectors).
The main challenge with many of these hand-crafted descriptors is that, they are not  able to scale well (in terms of accuracy and speed) on larger datasets.

Deep learning based models are used to overcome some of the limitations of the hand-crafted features.
Along this direction, convolutionahave been very successful in various computer vision and natural language processing tasks in recent years \cite{cnn1}-\cite{cnn10}.
Their success is mainly due to three factors: the availability of large-scale manually labeled datasets, powerful processing tools (such NVidia's GPUs), and good regularization techniques (such as dropout, etc) that can overcome overfitting problem.
They have been used for various problems in video processing too, such as action recognition \cite{act_cnn}, video classification \cite{c3d}, and video summarization \cite{videosum},  and significantly improved the performance over traditional approaches.
More interestingly, it is shown that the features learned from some of these deep architectures can be well-transferred to other tasks, i.e. one can get the features from a trained model for a specific task and use it for a different task (by training a classifier/predictor on top of it) \cite{offshelf}.

In this work we propose a deep learning framework for video commercial personalization, which has not been studied so far.
The previous approaches for commercial detection mostly rely on hand-crafted features derived from video \cite{com-det1}-\cite{com-det4}. 
This is a very important task, and of most interest for video broadcasters, content provides, and video sharing platforms (such as Youtube).
Video commercial personalization involves multiple sub-tasks, such as commercial segment detection, video content analysis, and commercial replacement based on user history.
Each one of these tasks requires to have a high-level understanding of the video semantic. 
Most of the previous deep learning based models for video analysis only rely on the visual content of the video. 
Although visual information is good enough for many of the video processing tasks, it is not sufficient for the video commercial analysis task.
This is mainly due to the visual similarity of many of the regular video contents and commercial videos, even to human eye. 
One such example is shown in Figure 1, where snapshots of two videos are provided, where in one Lionel Messi doing a commercial, and in the other one playing soccer.
As we can see these two images look very similar.
Usually only the last (and sometimes the first) few frames of the commercial videos are more relevant to the actual advertisement content, and the rest of the frames are not that correlated.
\begin{figure}[h]
\begin{center}
    \includegraphics [scale=0.18] {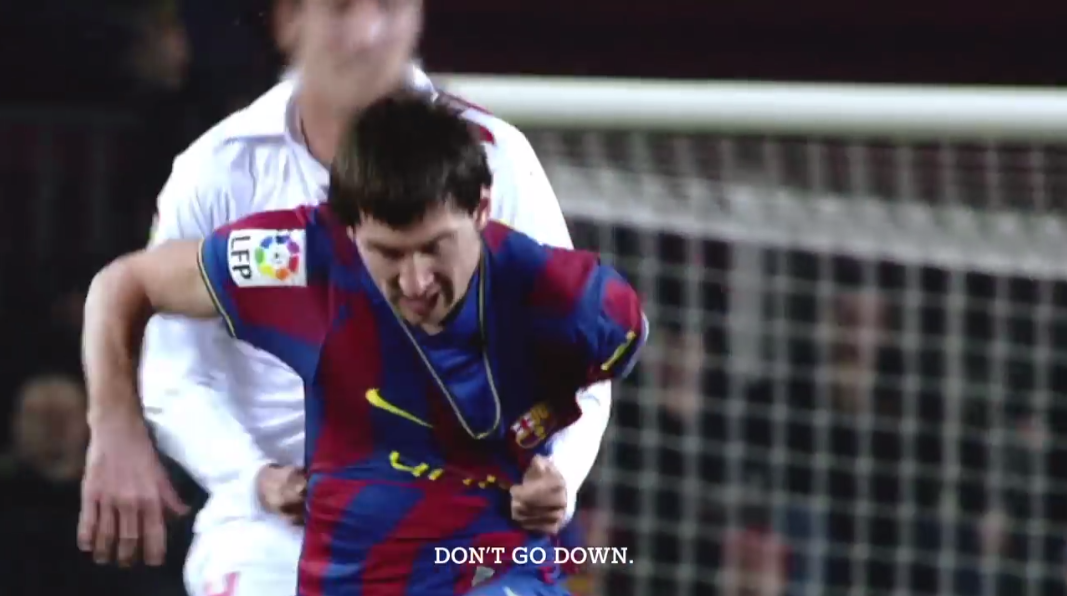}
    \includegraphics [scale=0.18] {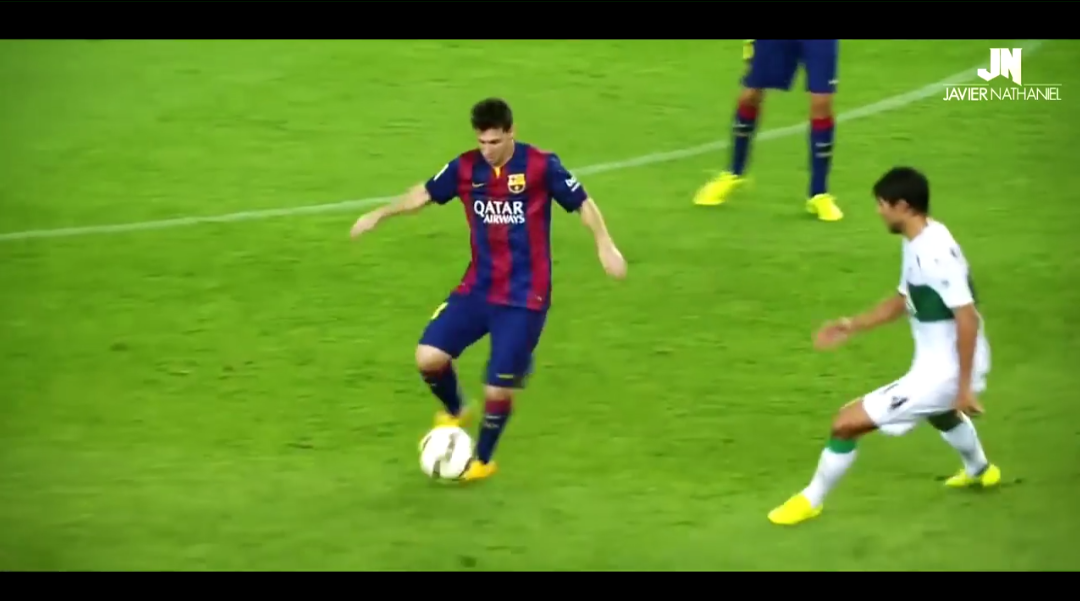}
\end{center}
  \caption{The snapshots of two videos, one from a commercial and the other from a sport game, courtesy of the content providers.}
\end{figure} 

To overcome the challenge with the visual similarity of videos and commercials, we propose a new two-stream convolutional network, which analyzes both the visual and audio information of the video to learn a representation and perform classification.
The model architecture is shown in Figure 2.
\begin{figure}[h]
\begin{center}
    \includegraphics [scale=0.34] {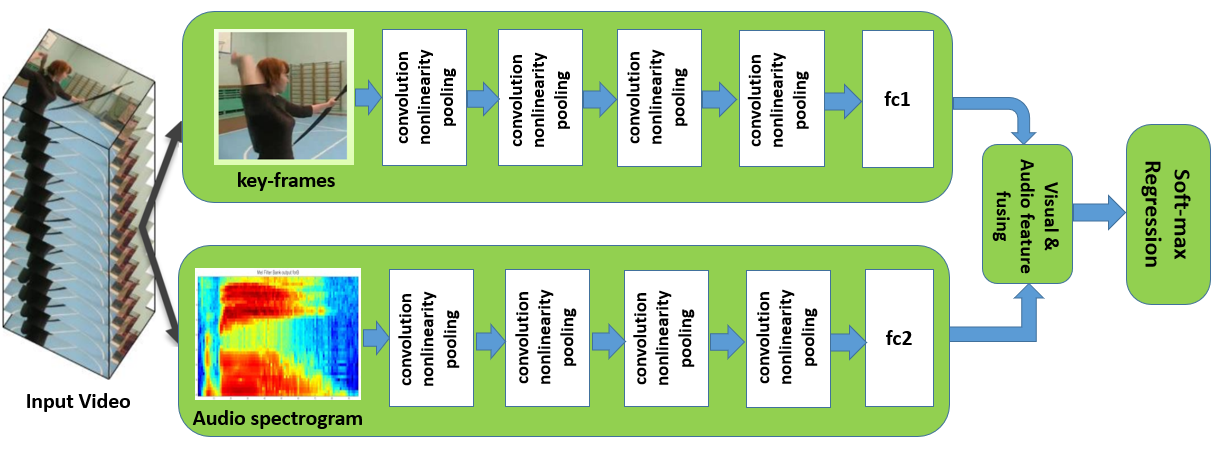}
\end{center}
  \caption{The block-diagram of the proposed audio-visual convolutional network}
\end{figure}

As we can see the top branch of this model processes the visual information, and the bottom branch analyzes the audio content of the video. 
Both visual and audio content go through multiple layers of abstraction to learn a suitable representation. 
The visual and audio representations are later fused together and are used for classification.
This model can be used for various video processing tasks, such as commercial detection, video classification, and video content analysis.
Here we study the application of this model mainly for commercial detection, which is a crucial task for personalized advertisement in broadcast videos (in which  the time-segment of the commercial is usually not available in the meta-data at the client side).
We collect a dataset of around 50k video and commercial shots. We then train this model on a subset of the dataset and evaluate on a test set. 
We show significant gain over traditional approaches, and also convolutional networks which only use visual content of the video.

The structure of the rest of this paper is as follows.
Section II provides the an introduction to the overall proposed framework, and briefly talks about different sub-tasks. 
Section III presents the detailed architecture of the proposed autio-visual convolutional network.
In Section IV, we provide the experimental studies and comparison with previous works.
And finally the paper is concluded in Section V.

\section{The Proposed Framework}
In this work we propose a machine learning framework for commercial detection and replacement in a more automatic and systematic way than previous approaches.
Most of previous approaches for commercial detection follow a multi-step process as shown in Figure 3.
One very important step in Fig. 3  is the feature extraction, which involves extracting various hand-crafted features from the video.
The discriminative power of these features largely determines the overall performance of these models.
\begin{figure}[h]
\begin{center}
    \includegraphics [scale=0.48] {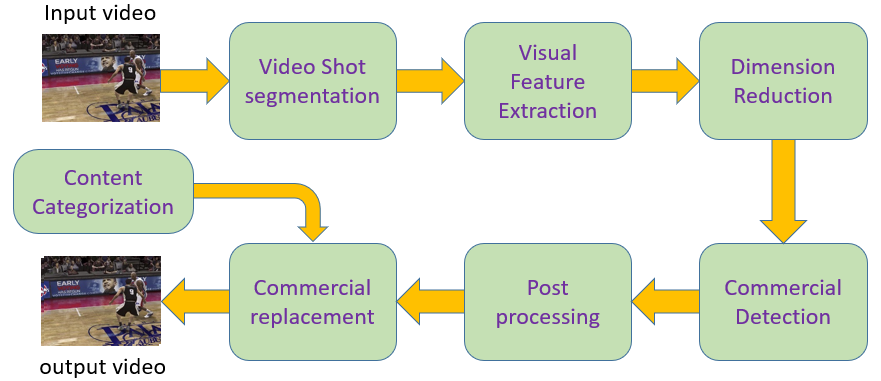}
\end{center}
  \caption{The block-diagram of commercial replacement framework}
\end{figure}

In our work, we try to skip the feature engineering step, and rely on a deep learning paradigm that jointly learns the video features, and performs commercial detection.
One main difference of our framework with previous deep learning approaches toward video processing, is that we rely on both the visual and audio information of the video. 
In this way, our model has the ability to learn a very general video representation that would suit for video commercial analysis.
The block diagram of our proposed model is shown in Figure 4.
\begin{figure}[h]
\begin{center}
    \includegraphics [scale=0.56] {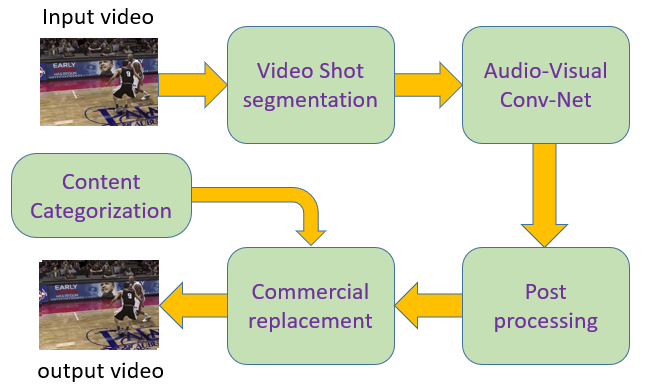}
\end{center}
  \caption{The block-diagram of proposed commercial detection model}
\end{figure}

We would like to mention that some of the steps are shared between our approach and those of previous works, such as video shot segmentation, and post-processing step. 
Video shot segmentation tries to segment the video temporally into shots of consistent scenes.
For shot segmentation, we use one of the available open sources \cite{shot-seg}. Then each video shot goes through a deep neural network that predicts if it is a commercial or regular video.
We then introduce a general model which can be used for both commercial detection, and video content analysis, and therefore can be trained in a multi-task learning fashion.
However, we put our main focus on commercial temporal segmentation within the video, which is a major step for personalized commercial recommendation.
We will provide the detailed explanation of the model architecture in the next section.

\section{Deep Audio-Visual Convolutional Network}
As discussed previously, we propose a deep neural network architecture for commercial detection, where it gets a segmented video shot as the input and classifies it as commercial or regular video content.
One can think of this as a binary classification problem.
Initially we only used the visual content of the video shot (video frames) as the network input (as shown in Figure 4) and it turns out it is not very easy to perform an accurate classification in that case.
We then designed a two-stream convolutional network which gets both the audio and visual information to perform the detection, and it turns out adding audio information to the network input significantly improves the detection accuracy.
One main reason for this is that regular video content and commercial can be very similar in terms of visual content, but if one can listen to the audio content of them, they would usually seem more different.
The detailed model architecture is shown in Figure 2.
Given a video shot, we extract a few key-frames from different parts of this shot and feed them as the input of the visual branch (top branch in Fig. 5). We also extract the audio content of the video and find its Spectrogram (which gives us a two-dimensional representation of the audio signal) and feed it as the input of the audio branch.
Then three layers of "convolution+nonlinearity+pooling" plus one fully connected layer are applied on each of these inputs to find an abstract representation of the signals.  
\begin{figure}[h]
\begin{center}
    \includegraphics [scale=0.38] {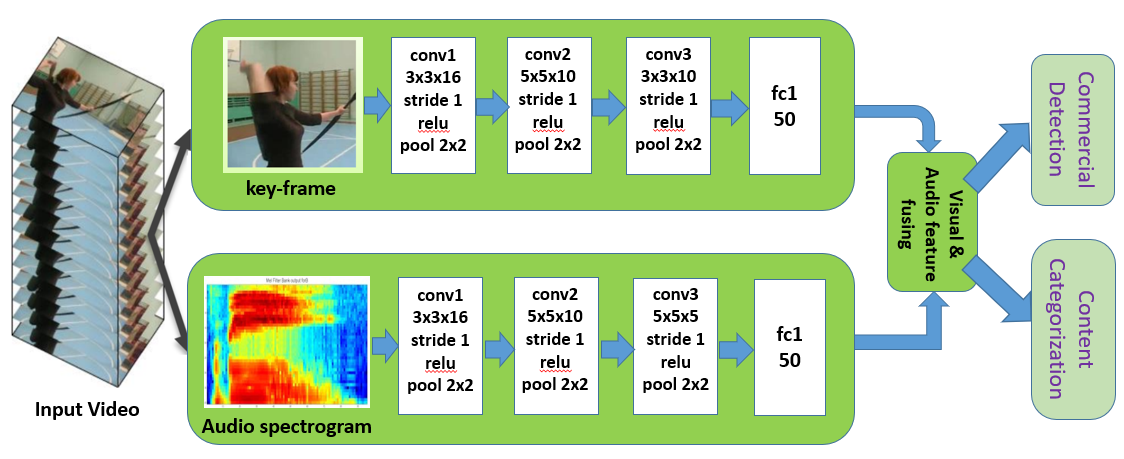}
\end{center}
  \caption{The block-diagram of the proposed audio-visual convolutional network}
\end{figure}

During training time, we also have a dropout layer before the last fully connected layer. 
We crop (the central portion) and resize each frame into a fixed size image of 112x112.
We also resample the audio part of the video such that all video shots have spectrogram of the same size.

To learn the parameters (weights and biases) of this model, we use a set of labeled video shots and minimize the prediction performance of the model.
Our loss function consists of two terms, one that measures the classification performance, and the other one imposes some $\ell_2$ regularization on the model weights (to avoid over-fitting).
\begin{equation}
\begin{aligned}
& \underset{W_i,b_i}{\text{min}}
& \sum_{i} y_i \log(y_i^{'}) +\lambda ( ||W_{fc1}^{(\text{visual})}||_F^2+ ||W_{fc1}^{(\text{audio})}||_F^2 )
\end{aligned}
\end{equation}
One can minimize this loss function using stochastic (more precisely mini-batch) gradient descent.
It is worth to mention that in case of multi-task learning where this model is used for both commercial detection and content analysis, we can add one more term (a cross-entropy) which evaluates the classification accuracy of content categorization part.
Therefore the overall loss function will be as the following equation in that case:
\begin{equation}
\begin{aligned}
& \mathcal{L}_{multi-task}=  \mathcal{L}_{Det}+   	\lambda_1 \mathcal{L}_{Cat}+ \lambda_2 \mathcal{L}_{Reg}
\end{aligned}
\end{equation}

\section{Experimental Results}
In this section we provide the experimental results for the proposed commercial detection model, and the comparison with the previous models based on hand-crafted features.
We also provide a comparison between audio-visual network performance, versus that of the visual network.
Our implementations are done in Python.
Tensorflow package \cite{TF} is used to train the proposed convolutional network.

Before providing the details of experiments, let us first talk briefly about our dataset.
We collected a dataset of around 51k regular video and commercial shots. 
There are about 27k regular video shots, and 24k commercial shots.
The regular video contents include movies, TV series, Sports, News, and documentaries (which covers the majority of contents in broadcast TVs).
The commercial shots are chosen from a variety of categories, which includes: foods, drinks, automobile, finance, etc.
The snapshots of 9 sample videos and commercial shots are shown in Figure 6.
\begin{figure}[h]
\begin{center}
    \includegraphics [scale=0.5] {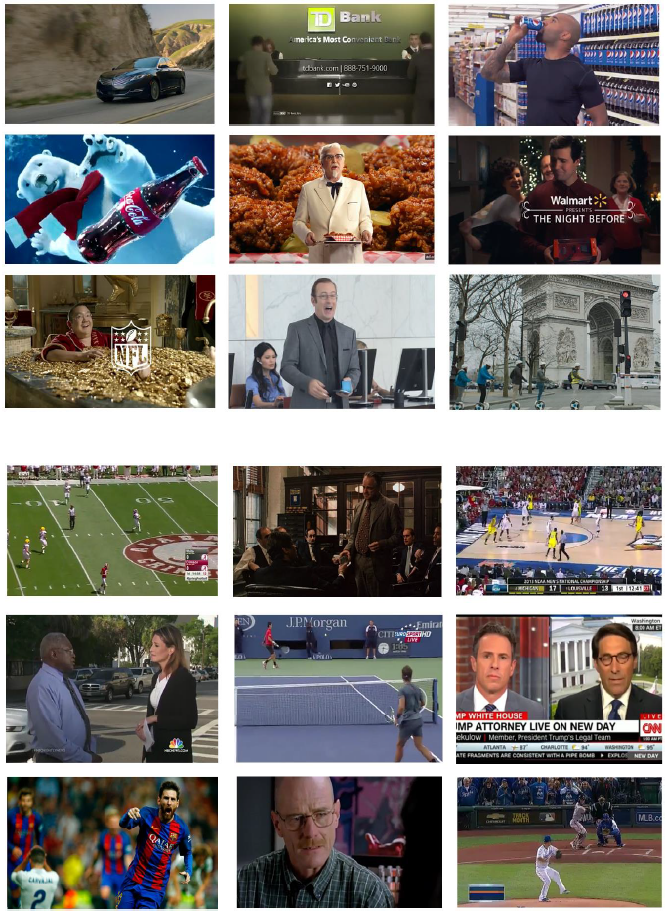}
\end{center}
  \caption{The images in top three rows denote the snapshots of 9 samples commercial shots, and the ones in the bottom three rows denote the snapshots of regular video content.}
\end{figure}

Now let us talk about our model hyper-parameters and the training procedure.
Stochastic gradient descent is used to train the model, with a batch size of 200.
The number of epochs (iterations over the entire dataset) is set to 100.
The learning rate is initialized with 0.001 and is decreased by factor of 0.95 after the 50-th epoch, until it reaches 0.00001 that is kept fixed after that.
To avoid over-fitting we apply dropout after the fully connected layers, with the probability of 0.5.
The weight for the $\ell_2$ loss is set to 0.0001, which is chosen based on a validation set.

We evaluate the performance of each model in terms of commercial detection accuracy. We also evaluate the precision and recall for some of the scenarios.

As the first experiment, we provide the comparison between the proposed model and some of the previous approaches based on hand-crafted features.
We compare with two hand-crafted features, histogram of oriented gradients (HOG) \cite{hog}, and local binary pattern (LBP) \cite{lbp}.
Table I provides a comparison between the accuracy of the proposed approach and the previous models.

\begin{table}[ht]
\centering
  \caption{Comparison of accuracy of different algorithms}
  \centering
\begin{tabular}{|m{3cm}|m{1.5cm}|}
\hline
Method  & Accuracy Rate\\
\hline
 LBP+SVM  &  \ \ \ \ \ \ 68.5\%\\
 \hline 
 HOG+SVM &   \ \ \ \ \ \ 76.4\% \\
\hline
 The proposed algorithm  &  \ \ \ \ \ \ 82\%\\
\hline
\end{tabular}
\label{TblComp}
\end{table}

Figure 7 present the training and test accuracy of the audio-visual convolutional network, with the visual convolutional network, and audio convolutional network, over different epochs.
\begin{figure}[h]
\begin{center}
    \includegraphics [scale=0.75] {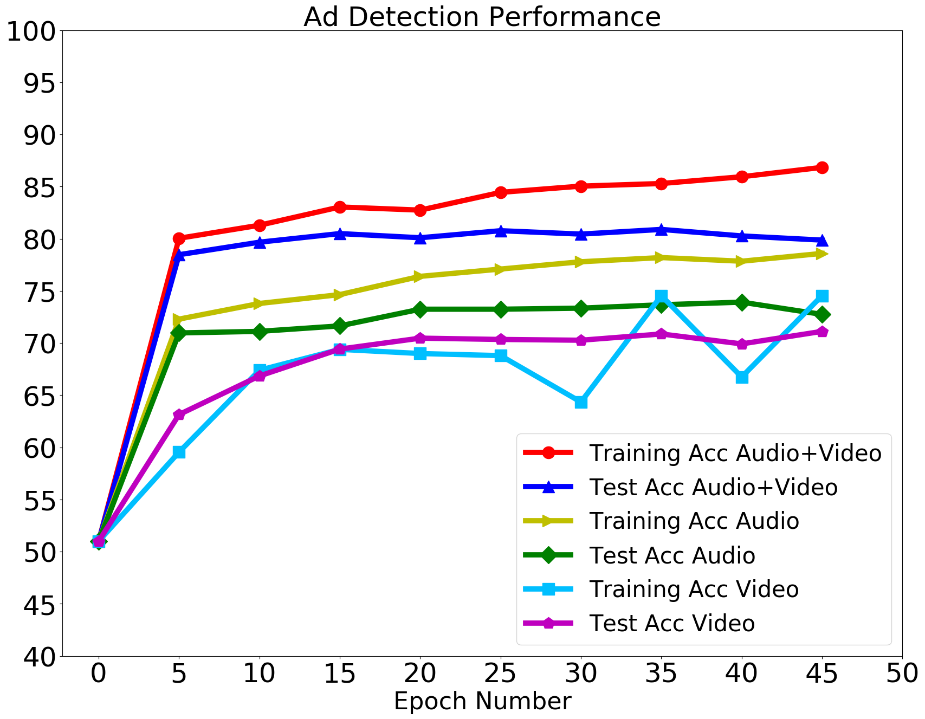}
\end{center}
  \caption{The training and test accuracies for different epochs}
\end{figure}

\section{Conclusion}
In this work we propose a commercial detection system using a two-stream convolutional network, where one branch processes the visual information of the video, and the other branch processes the audio content, and uses the fused audio-visual representation for detection.
Using the proposed model we provide a semi-automatic algorithm for commercial detection. 
Through experimental results, we show significant gain in detection accuracy over the previous approaches based on hand-crafted features.
The proposed convolutional architecture in this work can also be used for video content analysis, and commercial replacement, by training it in a multi-task learning fashion.

\section*{Acknowledgment}
We would like to thank Madhukar Budagavi, Youngkwon Lim, and Charlie Zhang for their suggestions and help during this project.
This work has been done while Shervin Minaee was doing a research internship at Samsung Research America.

\end{document}